# Collaborative Synthetic Data for Privacy-Preserving Financial Fraud Detection Across Organizational Silos

*Completed Research Paper*

**Simeon Allmendinger**
University of Bayreuth, AI Responsibility Centre, and Fraunhofer FIT, Universitätstraße 30, 95447 Bayreuth, Germany
Simeon.allmendinger@uni-bayreuth.de

**Domenique Zipperling**
University of Bayreuth, AI Responsibility Centre, and Fraunhofer FIT, Wittelsbacherring 10, 95444 Bayreuth, Germany
Domenique.zipperling@fit.fraunhofer.de

**Burhanettin Bahadir Kibar**
Technical University of Munich, Arcisstraße 21, 80333 Munich, Germany
burhan.kibar@tum.de

**Niklas Kühl**
University of Bayreuth, AI Responsibility Centre, and Fraunhofer FIT, Universitätstraße 30, 95447 Bayreuth, Germany
Kuehl@uni-bayreuth.de

## Abstract

*Organizations seek analytical value from AI, yet relevant data are often fragmented across organizations and constrained by privacy. This is acute in financial fraud detection, where rare fraud cases and imbalanced local datasets limit decision-relevant analytics. Federated learning enables collaboration without direct data sharing but does not resolve minority-class scarcity. Synthetic data generation can help, yet lightweight methods are interpolation-bound, while generative models require substantial data and computation. Existing collaborative generative approaches often rely on federated learning, imposing considerable organization-side training burdens. In this paper, we examine CollaFuse as a collaborative diffusion-based alternative for fraud detection and evaluate it across five fraud datasets. Compared with classical oversampling, local generative baselines, and centralized diffusion benchmarks, CollaFuse does not achieve the highest local fidelity but improves downstream fraud detection more consistently across most datasets. These findings suggest that synthetic data create analytical value less through local realism than through transferable cross-organizational structure.*

## Introduction

Sufficient and relevant data are critical organizational resources for the effective development of artificial intelligence (AI) systems (Jöhnk et al., 2021; Sharma et al., 2014). Yet in practice, data fragmentation across organizational silos and limited central accessibility hinder the mobilization of data for AI implementation and adoption (Li et al., 2025; Valtonen et al., 2026; Weber et al., 2023). While direct data sharing could mitigate these constraints, information systems (IS) research shows that the feasibility of such sharing depends not only on technical access, but also on privacy, governance, trust, data sovereignty, incentives, organizational readiness, and coordination mechanisms (Fassnacht et al., 2023; Finze et al., 2024; Hirt et al., 2025; Jussen et al., 2024; Opriel et al., 2025; Otto & Jarke, 2019). In response, collaborative learning methods such as federated learning (FL) have emerged to support inter-organizational analytics without requiring raw data sharing (McMahan et al., 2017). However, their utility is limited when the central constraint is not merely overall data scarcity, but the scarcity of informative minority-class observations, as is common in rare-event domains like fraud detection or healthcare (Sattarov & Schreyer, 2023).

Against this backdrop, synthetic data generation has emerged as an alternative to direct data sharing and as a potential means of mobilizing distributed data resources without exchanging raw records (Hirt & Kühl, 2018; Kaabachi et al., 2025; Li et al., 2025; Weber et al., 2023). Locally, lightweight methods, e.g., SMOTE (Chawla et al., 2002), efficiently address class imbalance but remain limited to interpolating observed minority cases and struggle with complex or heterogeneous distributions (Blagus & Lusa, 2013; Fernandez et al., 2018). More recent generative approaches, including generative adversarial networks (GANs) and diffusion models, promise higher-fidelity and more diverse samples (Hayaeian Shirvan et al., 2025; Khosravi et al., 2025; Pan et al., 2023), but they also require substantial data and computational resources. This creates a data dependency paradox: the very methods intended to overcome data scarcity depend on rich data to become effective. Recent work has therefore explored collaborative generative learning, including federated SMOTE, federated GANs, federated diffusion models, and synthetic data-sharing platforms for financial data (Bhutta & Mehmood, 2026; Karst et al., 2024; Sattarov et al., 2025; Sattarov & Schreyer, 2023). These approaches leverage distributed data without pooling raw records, but federated generative methods still require substantial local computation, as every organization must train the entire model, creating substantial computational and memory burdens. The inter-organizational challenge, therefore, remains: How can organizations collaborate to create high-fidelity synthetic data and analytical value when direct raw data sharing and local model training are both impractical?

Here, less common collaborative learning approaches, such as split learning (SL) (Gupta & Raskar, 2018), offer an alternative by partitioning model training between clients and a server, thereby reducing local computational requirements while preserving data locality. Recent work extends SL to denoising diffusion probabilistic models (DDPMs) (Ho et al., 2020), enabling collaborative training and high-fidelity synthetic data generation without sharing raw records (Allmendinger et al., 2026). In this study, we transfer this collaborative diffusion logic to tabular fraud detection, a setting in which class imbalance, heterogeneous institutional data, and data-sharing constraints make collaboration particularly relevant (Dal Pozzolo et al., 2014; Q. Huang et al., 2025; Yang et al., 2019). This allows us to examine whether collaborative generative learning can help organizations create useful synthetic minority-class data for downstream tasks when no single organization has sufficient local data and direct data sharing is constrained. We therefore ask (RQ): *To what extent can collaborative diffusion-based synthetic data generation create analytical value from distributed, imbalanced, and privacy-constrained data?*

To answer this question, we adapt collaborative diffusion-based generation from collaborative image synthesis to privacy-preserving synthetic fraud generation for imbalanced tabular classification. In our setting, multiple local clients—representing separate financial institutions—retain their raw transaction data locally, while a shared server supports collaborative synthetic data without accessing these raw records. We compare our approach against multiple alternatives, including different local interpolation-based oversampling methods (SMOTE, ADASYN), local generative tabular models (CTGAN, TabDDPM, local-only DDPM), and a centralized DDPM benchmark. To assess whether collaboratively generated synthetic data create analytical value, we evaluate them along two dimensions: their fidelity to fraud-related data distributions and their usefulness for downstream fraud detection across multiple classifier types.

Our results show that, in privacy-constrained inter-organizational settings, synthetic data need not closely reproduce local distributions to create value. Collaboratively generated samples improve fraud detection by transferring task-relevant cross-organizational patterns without pooling raw data. The contribution of this paper is therefore both methodological and conceptual. Methodologically, we adapt collaborative diffusion-based generation to imbalanced tabular fraud detection. Conceptually, we show that collaborative synthetic data generation can transform distributed, sensitive, and imbalanced data into decision-relevant analytical resources. This extends IS research on data-centric AI (Han et al., 2026; Jakubik et al., 2024), inter-organizational analytics, and privacy-preserving collaboration (Fassnacht et al., 2023; Jussen et al., 2024; Opriel et al., 2025; Otto & Jarke, 2019) by conceptualizing it as a governance-compatible mechanism for analytical value creation across organizational boundaries. The remainder of this paper reviews related work, introduces our collaborative diffusion-based approach, presents the experimental setup and results, discusses implications and limitations, and concludes with directions for future research.

## Background and Related Work

We position our study within a broader socio-technical challenge in IS: enabling inter-organizational analytics when valuable data are siloed and subject to privacy and governance constraints. Financial fraud detection is especially demanding because effective detection relies on rare minority-class cases unevenly distributed across institutions. Prior work offers partial solutions but has not established whether collaboratively generated synthetic data can transfer task-relevant structure across organizations in a governance-compatible way. **Table 1** summarizes prior work by indicating whether synthetic methods and classifiers are applied locally (Local) or inter-organizationally (Interorg.), whether they are empirically evaluated (Eval.), whether privacy concerns are addressed, and how many datasets are used for evaluation.

| Source | Synthetic method | | | Classifier | | | Privacy focus | # Data-sets |
|---|---|---|---|---|---|---|---|---|
| | Local | Interorg. | Eval. | Local | Interorg. | Eval. | | |
| Dal Pozzolo et al. (2014) | x | - | - | x | - | x | - | 1 |
| Strelcenia & Prakoonwit (2022) | x | - | - | x | - | x | - | 1 |
| Roy et al. (2024) | x | - | x | x | - | x | - | 2 |
| Karst et al. (2024) | x | - | - | - | x | x | x | 1 |
| Bhutta & Mehmood (2026) | - | x | - | - | x | x | x | 1 |
| Zhang et al. (2023) | x | - | - | - | x | x | x | 1 |
| Abadi et al. (2025) | - | - | - | - | x | x | x | 1 |
| Sattarov et al. (2023) | x | - | x | x | - | x | - | 3 |
| Sattarov & Schreyer (2023) | - | x | x | x | - | x | x | 2 |
| Sattarov et al. (2025) | - | x | x | x | - | x | x | 4 |
| Özcan et al. (2025) | x | - | - | - | x | x | x | 1 |
| Awosika et al. (2024) | - | - | - | - | x | x | x | 1 |
| J. Huang & Turetken (2025) | x | - | x | x | - | x | - | 1 |
| **Our study** | **x** | **x** | **x** | **x** | **x** | **x** | **x** | **5** |

**Table 1.** Related work on synthetic data and inter-organizational perspectives in fraud detection

### *Fraud Detection Under Class Imbalance*

Financial fraud detection has become increasingly important as the rapid digitalization of financial systems, driven by online banking, e-commerce, and mobile payment networks, has expanded the scale and speed of financial activity beyond the capacity of traditional manual or rule-based monitoring systems (Boulieris et al., 2024). Historically, financial fraud detection has served as a main defense mechanism for financial institutions (Bagwe, 2024), but static and expert-defined rules are difficult to scale and adapt to increasingly sophisticated fraud strategies that exploit complex behavioral patterns (Bhattacharyya et al., 2011). As a result, financial fraud detection has progressively shifted toward machine learning (ML) approaches, with prior studies examining a wide range of predictive models, including logistic regression, random forests, support vector machines, k-nearest neighbors, and neural networks (Afriyie et al., 2023; Baesens et al., 2021; Zhou et al., 2024). In ML-based fraud detection, the task is modeled as a binary classification problem that distinguishes fraudulent from legitimate transactions.

Let a dataset be denoted by $\mathcal{D} = \{(x_i, y_i)\}_{i=1}^{N}$, where $x_i \in \mathbb{R}^d$ represents the feature vector of transaction $i$, and $y_i \in \{0,1\}$ indicates whether the transaction is fraudulent. The goal is to learn a classifier that estimates the conditional probability $p(y = 1| x)$. Despite advances across model families, a persistent challenge in fraud detection is severe class imbalance, where fraudulent cases constitute only a small fraction of all transactions, often less than 1% (Van Belle et al., 2023). Consequently, standard ML algorithms tend to optimize overall accuracy by favoring the majority class, often at the expense of minority-class recall. In operational settings, however, failing to detect fraudulent cases is substantially more costly than misclassifying a limited number of legitimate transactions. This makes imbalance handling a central design requirement for fraud detection systems.

### *Traditional Method to Create Synthetic Data for Minority Fraud Cases*

A common way to address class imbalance is to augment the minority class with synthetic samples. Prior work often relies on interpolation-based oversampling, particularly SMOTE (Chawla et al., 2002) and ADASYN (He et al., 2008), which are widely used as classical baselines for imbalanced fraud detection (Dal Pozzolo et al., 2014; J. Huang & Turetken, 2025; Zhang et al., 2023). Both methods construct minority instances from local neighborhoods in feature space rather than by learning the underlying data distribution. SMOTE generates a synthetic minority sample by interpolating between an observed minority instance and one of its minority-class nearest neighbors. This procedure increases the density of the minority region while avoiding exact duplication of rare observations. However, SMOTE implicitly assumes that minority-class samples lie on a locally smooth and approximately convex manifold. This assumption can be problematic in fraud detection, where fraudulent behavior is often sparse, noisy, nonlinear, and multimodal. Because SMOTE generates new cases through local interpolation, it remains constrained by neighborhood geometry and may fail to capture irregular minority-class structure (Blagus & Lusa, 2013; Fernandez et al., 2018). ADASYN extends SMOTE by allocating more synthetic samples to minority observations that are harder to classify, but it inherits the same interpolation-based limitations and may further amplify noise in ambiguous boundary regions. Thus, although SMOTE and ADASYN remain important baselines for imbalance handling, they do not learn a global generative model of fraud. These limitations motivate the use of generative approaches that aim to model the underlying minority-class distribution more directly, a direction also reflected in recent work on synthetic financial data (Meldrum et al., 2025; Özcan et al., 2025; Roy et al., 2024; Sattarov et al., 2023; Strelcenia & Prakoonwit, 2022).

### *Generative Methods for Synthetic Minority Fraud Data*

Unlike interpolation-based oversampling methods, generative models aim to learn an approximation of the underlying data-generating distribution and generate new observations from that learned distribution. Two prominent families of generative models for synthetic data generation are GANs and diffusion models. GAN-based models rely on adversarial learning, where a generator produces synthetic observations and a discriminator attempts to distinguish them from real observations. Diffusion-based models follow a different principle. They learn to generate data by reversing a noising process. Because the collaborative extension examined in this study builds on diffusion models, we describe GAN-based approaches and provide a compact formalization of the DDPM objective as the basis for the methodological extension.

**GAN-based Generation.** GANs generate synthetic data through an adversarial training process in which a generator creates artificial samples and a discriminator learns to distinguish them from real observations (Goodfellow et al., 2020). For tabular data, standard GANs face challenges due to mixed numerical and categorical variables, non-Gaussian feature distributions, and imbalanced categories. CTGAN addresses these issues by adapting GANs to heterogeneous tabular data through conditional generation, mode-specific normalization, and training-by-sampling to better represent rare categories (Xu et al., 2019).

**Diffusion-based Generation.** Diffusion models provide a generative alternative to interpolation-based oversampling by learning the data distribution rather than constructing local convex combinations of observed points. In DDPMs (Ho et al., 2020), a forward process gradually corrupts a data sample $x_0$ by adding Gaussian noise over $T$ timesteps. The Markov transition from step $t-1$ to $t$ is defined as

$$q(x_t \mid x_{t-1}) = \mathcal{N}\left(x_t; \sqrt{1-\beta_t}\, x_{t-1}, \beta_t I\right),$$

where $\beta_t \in (0,1)$ is a predefined variance schedule. Equivalently, a noisy sample at timestep t is given by

$$x_t = \sqrt{\overline{\alpha_t}}\; x_0 + \sqrt{1-\overline{\alpha_t}}\,\epsilon, \qquad \epsilon \sim \mathcal{N}(0, I), \qquad \text{with} \qquad \alpha_t = 1-\beta_t, \qquad \overline{\alpha_t} = \prod_{s=1}^{t} \alpha_s.$$

A neural network $\epsilon_\theta(x_t, t)$ is then trained to predict the added noise $\epsilon$, thereby learning the reverse denoising process. A standard simplified training objective is

$$\mathcal{L}_{\text{diff}}(\theta) = E_{x_0,\epsilon,t}[\|\epsilon - \epsilon_\theta(x_t, t)\|_2^2].$$

Diffusion models are not limited to local interpolation between existing minority-class observations. Instead, they learn a generative process that can capture broader distributional patterns, making them relevant for complex and heterogeneous fraud data. TabDDPM extends this principle to tabular settings by jointly modeling numerical and categorical features and has shown strong performance on tabular benchmarks (Kotelnikov et al., 2023). However, both CTGAN and TabDDPM are typically designed under centralized training assumptions, where all data are available in a single location. In inter-organizational and privacy-sensitive settings, this assumption is unrealistic as transaction data usually cannot be pooled across organizations due to privacy, regulatory, or organizational constraints (Q. Huang et al., 2025; Yang et al., 2019). This creates a gap between advances in tabular synthetic data generation and the collaborative settings in which fraud detection is often deployed, motivating privacy-preserving collaborative diffusion learning as a way to benefit from distributed fraud data without requiring direct data sharing.

## *Privacy-Preserving Collaborative Diffusion Learning*

The centralized training assumption discussed in the previous section is difficult to satisfy in financial fraud detection, where transaction data are distributed across institutions and cannot be freely pooled due to privacy, regulatory, or organizational constraints. This creates an important organizational problem: fraud patterns may span multiple banks, payment providers, or platforms, while each institution observes only a partial view of the underlying activity. Effective fraud detection, therefore, depends on inter-organizational collaboration, yet IS research shows that data sharing across organizations is shaped not only by technical feasibility but also by trust, data sovereignty, governance, incentives, organizational readiness, and coordination mechanisms (Fassnacht et al., 2023; Jussen et al., 2024; Opriel et al., 2025; Otto & Jarke, 2019). To address such data silos, prior work has studied collaborative learning paradigms such as FL and SL. FL enables clients to jointly optimize a global model while keeping their data local (McMahan et al., 2017) and has also been explored in fraud detection settings (Awosika et al., 2024; Q. Huang et al., 2025; Yang et al., 2019). In this setting, organizations train local model copies on their data and share parameter or gradient updates with a central server, which aggregates them into a global model. A drawback of FL is that each organization must train a full model. An alternative is SL, which reduces local computation by partitioning the model between organizations and a server (Gupta & Raskar, 2018). In SL, the organization computes only the initial layers of the model and sends intermediate activations to the server, which completes the remaining forward and backward computations. Both approaches are shown in **Figure 1**.

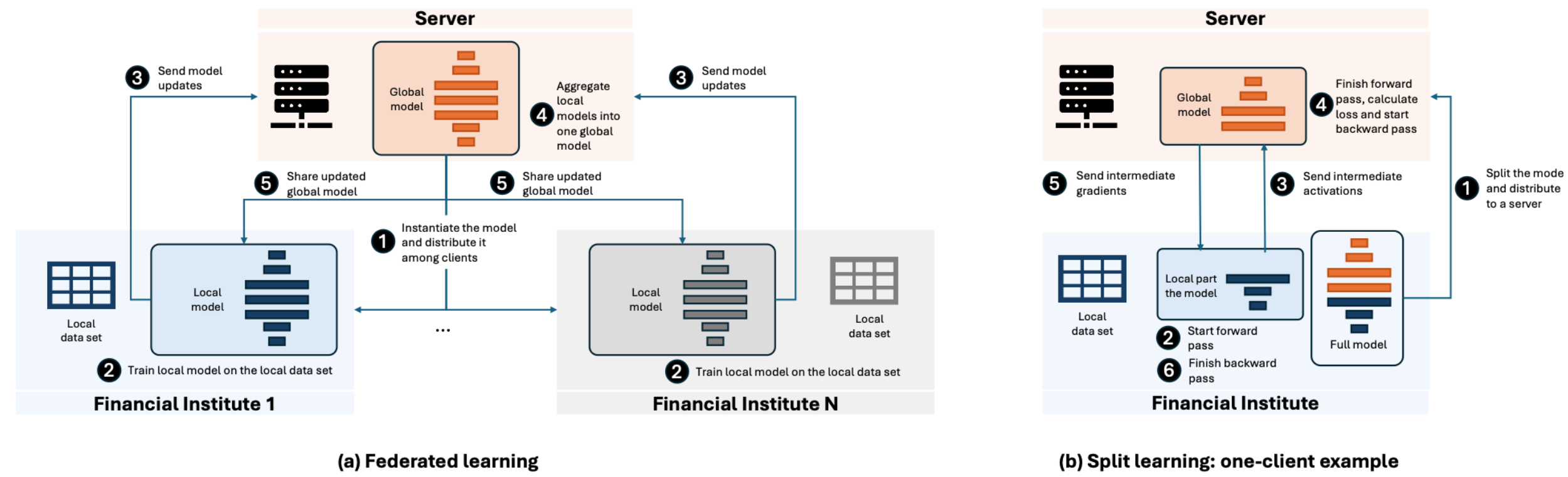

**Figure 1**. Illustration of federated learning and one-client split learning

Building on this collaborative-learning logic, recent work has begun to examine synthetic data generation as a privacy-preserving mechanism for fraud and financial tabular data. Simple approaches involve the sharing of locally generated synthetic data (Karst et al., 2024). Additional approaches have been explored, including federated SMOTE, federated GAN- and federated diffusion-based generation for imbalanced credit-card fraud detection (Bhutta & Mehmood, 2026), as well as FedTabDiff, which adapts diffusion probabilistic models to federated mixed-type tabular data generation (Sattarov et al., 2025; Sattarov & Schreyer, 2023). These studies demonstrate the growing relevance of collaborative synthetic data generation in privacy-sensitive financial settings. However, these approaches focus on federated settings in which organizations must still train the generative model locally, thereby creating substantial organization-side computational and memory burdens for collaborative minority fraud generation. This limitation motivates the use of collaborative diffusion-based generation, as proposed by Allmendinger et al. (2026) for the generation of synthetic fraud cases. This collaborative diffusion approach builds on the logic of SL and extends it from network layers to diffusion timesteps. Instead of partitioning a neural network into organization-side and server-side layers, it splits the denoising trajectory at a timestep cut point $\tau_{\text{cut}}$. The server performs the earlier part of the reverse diffusion process, while each organization performs the final denoising steps locally. The underlying intuition is that early denoising steps operate on noisy representations and are therefore less privacy-sensitive, whereas later denoising steps reconstruct finer-grained, potentially identifying structure and should remain on the client side.

Formally, let $T$ denote the total number of timesteps and $\tau_{\text{cut}} \in \{1, \dots, T\}$ the split point. Then the generative process is partitioned into a server-side component for timesteps $t > \tau_{\text{cut}}$ and an organization-side component for timesteps $t \leq \tau_{\text{cut}}$. The collaborative distribution is expressed as

$$p_{\theta_s,\theta_c}(x_{0:T}) = p(x_T) \prod_{t=\tau_{\text{cut}}+1}^{T} p_{\theta_s}(x_{t-1} \mid x_t) \prod_{t=1}^{\tau_{\text{cut}}} p_{\theta_c}(x_{t-1} \mid x_t),$$

where $\theta_s$ are the server parameters, and $\theta_c$ are the parameters of organization $c$. This formulation allows organizations to outsource computationally expensive denoising steps while retaining local control over the final, more privacy-sensitive reconstruction stages. The resulting reduction in organization-side effort can be made concrete. As each reverse-diffusion step requires a full forward pass through a denoising network, organization-side generation cost scales approximately linearly with the share of timesteps executed locally. At the cut points reported as most effective, clients execute at most 20% of denoising steps, such that more than 80% of the iterative generation workload is delegated to the shared server. Moreover, in contrast to federated generative approaches, in which every organization trains the complete generative model locally and exchanges full model parameters in each communication round (Allmendinger et al., 2026; McMahan et al., 2017), the timestep-split design exchanges only noised representations, and the organization-side model must approximate only the final segment of the denoising trajectory.

Allmendinger et al. (2026) introduced and evaluated this collaborative diffusion architecture in the context of image synthesis, showing that collaborative diffusion can improve generation quality relative to isolated organization models while enabling a privacy-performance trade-off through the choice of $\tau_{\text{cut}}$. However, its original formulation is developed for image data and does not address the specific characteristics of financial fraud detection. Tabular fraud data differ from images as they contain mixed numerical and categorical variables, severe class imbalance, sparse minority patterns, heterogeneous institutional distributions, and domain-specific validity constraints. As a result, it remains unclear whether this timestep-split collaborative diffusion principle can be transferred to tabular fraud data and whether the resulting synthetic minority samples improve downstream fraud detection.

Beyond this technical question, prior work points to a broader inter-organizational IS challenge: how organizations can create analytical value from rare, sensitive, fragmented, and computationally constrained data without pooling raw records (McMahan et al., 2017). Local methods preserve organizational autonomy but remain limited by scarce minority-class observations (Dal Pozzolo et al., 2014); federated generative approaches mobilize distributed data but require substantial client-side training and memory resources (Bhutta & Mehmood, 2026; Sattarov et al., 2025; Sattarov & Schreyer, 2023), and centralized approaches offer useful bound benchmarks but are often infeasible under privacy and governance constraints. Yet existing work has not systematically compared these strategies across both synthetic data fidelity and downstream task performance. We address this gap by evaluating timestep-split collaborative diffusion as a governance-compatible and computationally feasible mechanism for inter-organizational analytical value creation in and beyond fraud detection. Thereby, our study aims to inform the design and selection of synthetic data strategies for privacy-constrained inter-organizational analytics and value creation.

## Collaborative Diffusion for Privacy-Preserving Fraud Detection

This section presents our adaptation of timestep-split collaborative diffusion to collaborative tabular fraud detection and the corresponding experimental design. Across five fraud datasets, we implement a reproducible three-stage pipeline: collaborative training of the diffusion model, local generation of synthetic minority-class fraud samples, and downstream fraud detection using augmented local datasets. We evaluate the approach along two dimensions—synthetic data fidelity and downstream classification utility—and compare it against real-data-only, oversampling, and alternative generative baselines.

### *Training a Collaborative Diffusion Model*

Building on timestep-split collaborative diffusion, we instantiate collaborative diffusion for the generation of synthetic minority-class fraud samples in an inter-organizational tabular setting. The central adaptation is a shift in application objective: whereas prior work applies timestep-split collaborative diffusion to image synthesis, our setting uses it as a privacy-preserving augmentation mechanism for imbalanced fraud detection. Accordingly, the method is not evaluated by perceptual sample quality, but by its ability to generate useful synthetic fraud cases for downstream classification under distributed data constraints. From Allmendinger et al. (2026), we adopt the timestep-split partition of the reverse diffusion process at a cut point, the separation into a shared server model and organization-side models without shared weights exchanging only noised intermediate representations instead of raw records. We newly adapt the generative backbone to mixed-type tabular transaction data, and shift evaluation from perceptual fidelity to downstream detection utility.

Each organization $c \in \{1, \ldots, K\}$ holds a private local dataset $\mathcal{D}_c = \{(x_i, y_i)\}_{i=1}^{n_c}$, where $x_i \in R^d$ denotes a tabular transaction vector and $y_i \in \{0,1\}$ indicates whether the transaction is fraudulent. The collaborative objective is to learn a generative process that captures globally useful fraud structure without exposing raw records. Following this timestep-split design, training is therefore split across diffusion timesteps: the server model learns from noisy latent states in later denoising steps, while the organizational model retains the local and more privacy-sensitive part of the denoising trajectory. In the fraud setting, this design serves two purposes simultaneously: it preserves data locality and enables synthetic augmentation beyond purely local interpolation. To operationalize the split, the cut point is defined as $\tau_{\text{cut}} = \lfloor \tau_{\text{ratio}} \times T \rfloor$, where $T$ is the total number of diffusion timesteps and $\tau_{ratio} \in [0,1]$ controls how much of the reverse process is handled locally versus by the server. During training, each organization computes the forward process up to the cut point and transmits only the noised representation $x_{\tau_{\text{cut}}}$ rather than the original transaction record $x_0$. The server then optimizes the denoising objective for the later timesteps and thereby learns shared fraud-relevant structure from deliberately corrupted intermediate states. This setup follows the core logic of timestep-split collaborative diffusion but is instantiated here for tabular fraud generation.

Our privacy argument rests on architectural separation. We assume an honest-but-curious server that follows the protocol but may attempt to infer information from the data it receives, and non-colluding organizations that do not exchange raw records. Thus, during training, the server receives only representations noised up to the cut point, and during generation, it operates on samples drawn from pure Gaussian noise. Three residual leakage channels remain. First, the noised representations disclosed during training still carry a residual signal about the underlying records. This signal shrinks as the cut point increases, and empirical analyses show that attribute-inference and reconstruction attacks degrade at higher noise levels (Allmendinger et al., 2026). Second, synthetic fraud samples may memorize rare training records, a known risk for generative models that is amplified by small minority classes (Stadler et al., 2022). Third, a malicious participant could probe the shared server model to approximate fraud structure contributed by other organizations. Formal mechanisms such as differentially private training could be layered onto the architecture at additional utility cost. We do not claim such guarantees here.

A second adaptation concerns the training objective. In contrast to a standard tabular diffusion model trained only with denoising loss, our implementation uses a fraud-aware composite objective similar to Roy et al. (2024) to make the generated samples more useful for minority-class augmentation. The total organization-side loss is defined as

$$\mathcal{L}_{total} = \mathcal{L}_{norm} + w_1 \mathcal{L}_{prior} + w_2 \mathcal{L}_{triplet}.$$

Here, $\mathcal{L}_{\text{norm}} = \|\epsilon - \epsilon_\theta(x_t, t)\|_2^2$ is the standard reconstruction term. The prior regularization term $\mathcal{L}_{prior}$ encourages the predicted denoising errors to remain statistically consistent with the non-fraud distribution, thereby reducing drift in generated tabular values. The triplet term $\mathcal{L}_{triplet} = \max(0,\ d(\hat{x}, p) - d(\hat{x}, q) + \gamma)$, where $\hat{x}$ denotes a generated fraud sample, $p$ a real fraud observation (positive sample), $q$ a real non-fraud observation (negative sample), and $\gamma > 0$ the margin. This loss encourages generated fraud samples to lie closer to real fraud observations than to real non-fraud observations, thereby counteracting collapse toward the majority class and aligning generation with the downstream detection objective.

After training, each organization generates synthetic fraud cases by sampling Gaussian noise and applying the collaborative reverse process. The server performs denoising from timestep $T$ down to $\tau_{\text{cut}}$, and the organization completes the final local denoising steps to obtain a synthetic fraud sample $\widehat{x_0}$. These generated fraud samples are not pooled globally; they are assigned back to the originating organization and used only for organization-level augmentation. This preserves the institutional separation assumed throughout the study and keeps the method aligned with privacy-preserving scenarios. In subsequent steps, this synthetic dataset can be used to augment the real-world data and utilized to train a classifier locally.

**Figure 2** illustrates the generation steps of new synthetic samples following the collaborative training of the DDPM, which are subsequently used to train the classifier: (1) an organization (here a financial institution) samples Gaussian noise individually, (2) sends it to the server, (3) the server partially denoises the data, (4) sends the partial results back to the respective organization, (5) the organization finishes the generation process with its local part of the DDPM and (6) uses it together with the real-world data to train a classifier. For further details on the training scheme and algorithm used for collaborative DDPM training, we refer the reader to Allmendinger et al. (2026).

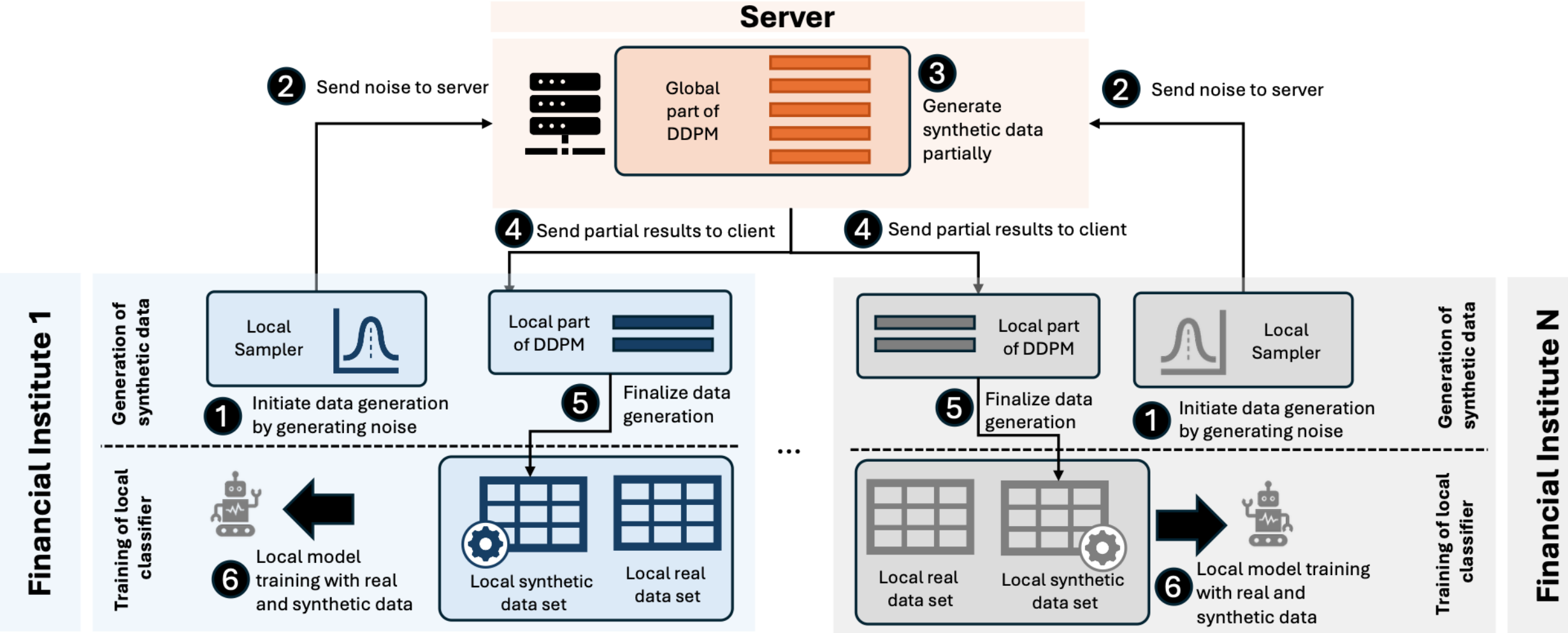

**Figure 2**. Schematic application of collaborative diffusion for privacy-preserving fraud detection after collaborative training of the DDPM.

## *Experimental Setup*

We implement the experiments as a reproducible pipeline across five fraud datasets to avoid relying on a single empirical setting. For each dataset, observations are first transformed into an organization-level format that reflects the intended inter-organizational setting and enables consistent comparison between local, collaborative, and centralized training scenarios.

We organize the empirical pipeline as a three-stage process. In Stage 1, organizations collaboratively train a timestep-split diffusion model while keeping raw transaction data local. In Stage 2, the trained model is used to generate organization-specific synthetic minority-class fraud samples, which are compared with local oversampling and generative baselines in terms of distributional fidelity (see steps 1 to 5 in **Figure 2**). In Stage 3, these synthetic samples are used to augment local training data for downstream fraud detection, and classifier performance is evaluated against benchmarks using only real data and alternative augmentation methods (see step 6 in **Figure 2**).

### Stage 1: Collaborative Training of the Diffusion Model

Organization construction follows dataset-specific rules designed to induce realistic heterogeneity across local data holders. For the IEEE-CIS dataset (Howard & Bouchon-Meunier, 2019), for example, organizations are defined using the payment-card attributes card4 and card6, which capture card-network and card-type characteristics and therefore provide a meaningful proxy for institution-like transaction groups. For PaySim, we partition on the categorical transaction type, as three of the five types contain no fraud cases. The corresponding organizations cannot support local training and are excluded, leaving two organizations. For BAF, Elliptic, and Credit Card Fraud, we apply quantile splits over each dataset's temporal index, so each pseudo-organization corresponds to a distinct temporal cohort. This creates non-identical organization distributions and reflects the intended inter-organizational setting, in which transaction patterns vary across organizations and raw records remain decentralized throughout model training, synthetic data generation, and downstream evaluation. The induced heterogeneity is substantial. Per-organization fraud prevalence varies by up to 9.2× (Elliptic, 1.8%–16.2%) and local sample sizes by up to 14× (IEEE-CIS). Each local dataset is further divided into a training set and a test set, with the training set used for collaborative model training and later augmentation, and the test set reserved for fidelity and downstream evaluation.

**Synthetic Data Generation and Augmentation Methods.** The empirical comparison of the synthetic data's fidelity includes generation methods alongside traditional benchmarks such as reweighting and oversampling. Specifically, we compare random oversampling, SMOTE, ADASYN, our approach, and alternative generative tabular models, including CTGAN (Xu et al., 2019) and TabDDPM (Kotelnikov et al., 2023). We also include a centralized DDPM trained on pooled data as a benchmark for a theoretical scenario in which all data are shared. It is important to note that we oversampled or trained the generative models only on fraud cases. We also only generated synthetic fraud cases.

**Collaborative Training.** In the first stage, collaborative diffusion is trained collaboratively across organizations using our tabular extension. Organizations' raw data remain local throughout this process. Instead of sharing transaction records, organizations participate in joint diffusion training through the exchange of noisy intermediate representations, consistent with the collaborative diffusion design. This stage establishes the shared generative model that is later used to create synthetic fraud samples.

### Stage 2: Local Generation of Synthetic Fraud Samples

After collaborative training, the trained diffusion model is used to generate organization-specific synthetic fraud samples locally. To benchmark this generation stage, we compare our approach against both classical oversampling and alternative tabular generative baselines. Specifically, the comparison includes random oversampling, SMOTE, ADASYN, CTGAN, TabDDPM, local-only DDPM, and a centralized DDPM trained on pooled fraud cases as a reference benchmark for a theoretical full-data-sharing scenario. Across all methods, we restrict training and generation to the fraud cases only, such that all synthetic observations correspond exclusively to the minority class.

Random oversampling, SMOTE, and ADASYN are applied locally to each organization's fraud cases in the training partition. For SMOTE and ADASYN, the neighborhood size is set to five. CTGAN, TabDDPM, and local-only DDPM are trained separately on each client's local fraud cases. The centralized DDPM is trained on pooled fraud cases across clients and serves as a non-privacy-preserving upper-bound benchmark under centralized data access. In contrast, collaborative diffusion is trained collaboratively across organizations while preserving data locality and is then used to generate organization-specific fraud samples.

**Fidelity Evaluation of Synthetic Fraud Samples.** The second stage evaluates the distributional fidelity of the generated synthetic fraud samples. For this purpose, the pipeline produces dedicated fidelity artifacts, including Maximum Mean Discrepancy summaries, embedding files, and visualization plots. Qualitatively, real and synthetic fraud samples are projected into a low-dimensional representation using PCA followed by t-SNE (see **Figure 3** (a-e)). Quantitatively, we compute Maximum Mean Discrepancy (MMD) between real and synthetic client-level fraud samples. In the default configuration, PCA is reduced to ten components before visualization, and MMD is estimated across multiple random seeds to account for stochasticity in the generation process. Lower MMD values indicate greater statistical proximity between the synthetic and empirical client-level fraud distributions.

### Stage 3: Downstream Fraud Detection and Evaluation

The third stage examines the downstream utility of the generated synthetic data in a supervised fraud detection task. For each organization, classifiers are trained either on the original local training data or on locally augmented training data that include additional synthetic fraud samples generated in Stage 2. Real-only training, both with and without class weighting, serves as the benchmark. The classifier suite comprises Logistic Regression, Random Forest, LightGBM, and HistGradientBoosting, thereby covering linear, bagged-tree, and boosting-based learners commonly used in fraud detection. In addition, augmentation strength is varied through a ratio grid from 0.1 to 0.5, which controls the number of injected synthetic fraud samples relative to the residual local class imbalance. We additionally evaluated a FL-based benchmark by training logistic regression via Federated Averaging. As this benchmark did not produce competitive downstream results compared with the augmentation-based alternatives, we exclude it from the focal comparison reported below.

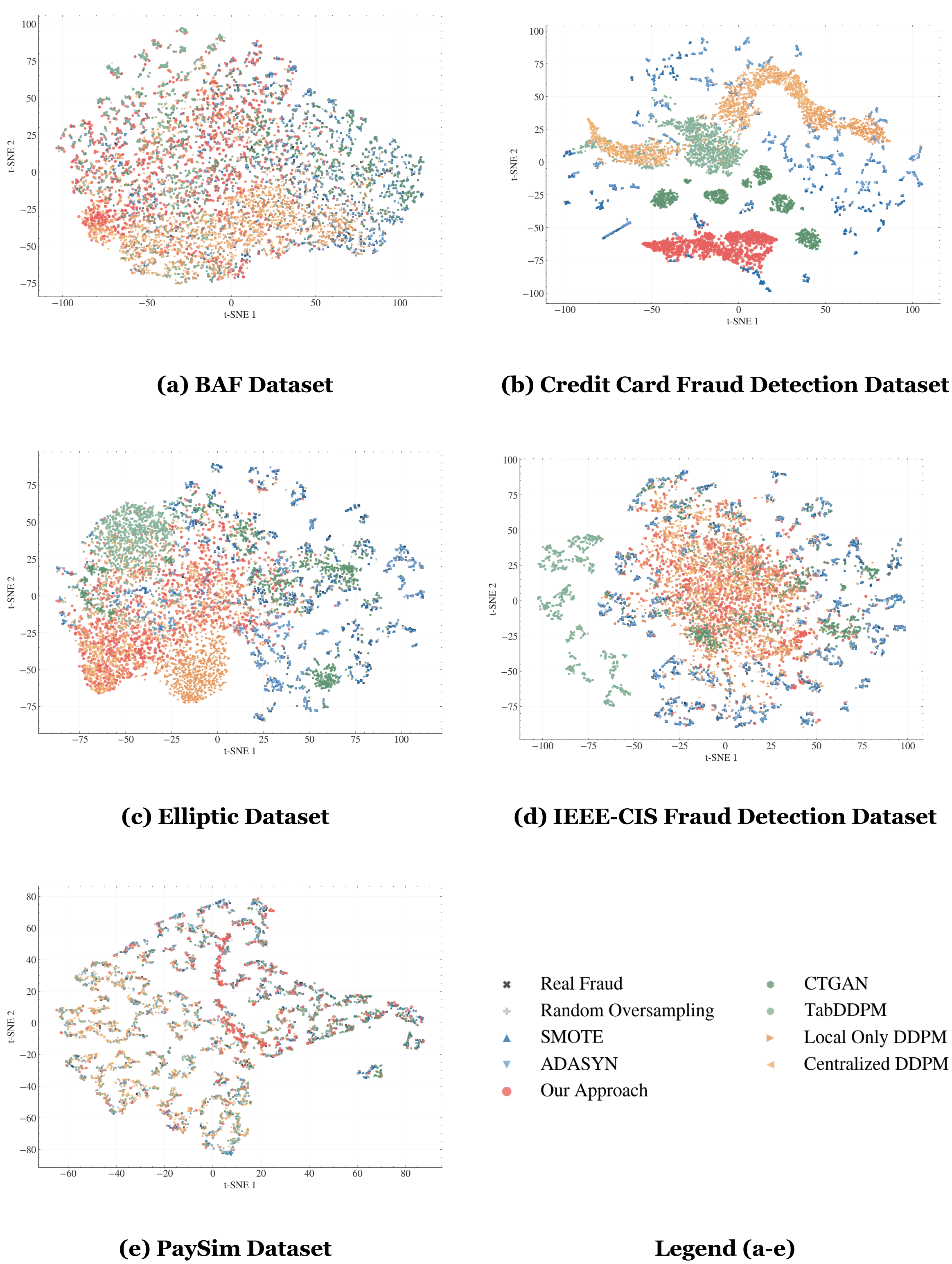

**(a) BAF Dataset**

**(b) Credit Card Fraud Detection Dataset**

**(c) Elliptic Dataset**

**(d) IEEE-CIS Fraud Detection Dataset**

**(e) PaySim Dataset**

**Legend (a-e)**

**Figure 3.** PCA+t-SNE projection of real and synthetic fraud samples by method. Greater overlap indicates better distributional alignment.

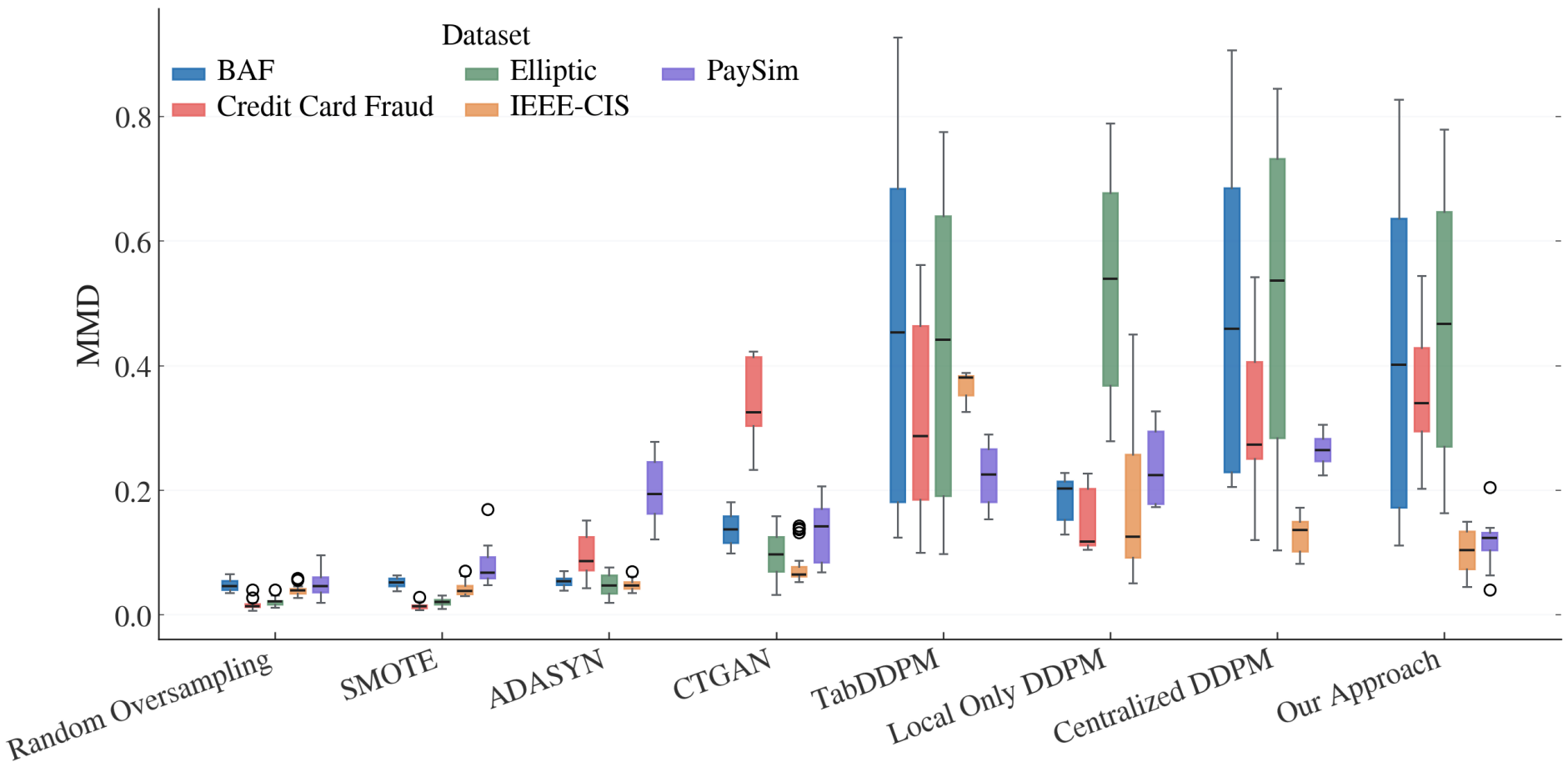

**Figure 4.** Client-level fidelity across datasets. Boxplots show MMD↓ between real and synthetic fraud samples; lower values indicate closer alignment with the local fraud distribution.

***Performance Metrics and Aggregation.*** Downstream performance is evaluated on untouched real organization-specific test sets. We report F1-score as the primary performance metric. Aggregate comparisons across organizations, augmentation ratios, augmentation sources, and classifier families are then used to assess whether collaborative diffusion provides systematic advantages over classical local resampling, local generative baselines, and real-data-only training. The implementation is available online (https://github.com/SimeonAllmendinger/collafuse-for-fraud-detection).

***Datasets.*** To avoid concluding from a single fraud benchmark, we evaluate the pipeline on five datasets that differ in feature structure and organizational heterogeneity. The current implementation supports IEEE-CIS Fraud Detection (Howard & Bouchon-Meunier, 2019), Bank Account Fraud (BAF) (Jesus et al., 2022), PaySim (Lopez-Rojas et al., 2016), Credit Card Fraud Detection (*Credit Card Fraud Detection*, n.d.), and the Elliptic Data Set (Weber et al., 2019). In all cases, the preparation stage converts the raw data into a common tabular representation and creates one aligned training and test split per organization, enabling a consistent comparison of augmentation methods across otherwise heterogeneous fraud settings.

The datasets were selected because they expose different empirical challenges. IEEE-CIS is a large-scale e-commerce fraud dataset with separate transaction and identity tables joined by TransactionID, making it comparatively rich in mixed transactional and identity-related features. Credit Card Fraud Detection is a highly imbalanced card-transaction benchmark with PCA-transformed variables, which makes it cleaner structurally but less interpretable feature-wise. PaySim differs from both by representing simulated mobile money transactions and therefore emphasizes transaction-type dynamics rather than card or identity information. BAF adds another perspective by providing privacy-preserving, realistic tabular data for bank account opening fraud and explicitly captures temporal dynamics and strong imbalance. Finally, Elliptic is distinct from the others because it originates from anti-money-laundering in Bitcoin transactions.

A second source of variation lies in how organizations are constructed. For IEEE-CIS, organizations are defined using observed card4 and card6 combinations, yielding subpopulations that resemble payment-network and card-type segments. PaySim uses transaction type as the organization split source, which creates behaviorally distinct transaction groups. BAF, Credit Card Fraud Detection, and Elliptic instead use deterministic quantile-based pseudo-organizations, which allow the evaluation to simulate decentralized institutions even when no natural organization identifier exists in the raw data.

## Results

This section reports the empirical findings of our evaluation. Following the study design, we assess the proposed approach along two dimensions: (1) the fidelity of generated fraud samples and (2) their downstream utility for fraud detection. The evaluation compares our approach against classical resampling baselines and alternative generative tabular baselines across multiple datasets and client partitions.

**Distributional Fidelity of Synthetic Fraud Samples.** We first examine whether collaborative diffusion generates synthetic fraud samples that are statistically close to the empirical organization-level fraud distributions. To this end, we evaluate sample fidelity qualitatively through PCA+t-SNE projections and quantitatively through Maximum Mean Discrepancy (MMD) displayed in **Figure 4**. Lower MMD values indicate greater similarity between real and synthetic data distributions. Across datasets, the results show a clear pattern: local interpolation-based methods generally achieve the lowest MMD values, while our approach exhibits higher distributional distance from the organization-level distributions. The same tendency holds relative to other generative baselines, whose fidelity varies by dataset and organization composition. Overall, these findings suggest that methods based on local geometry remain strongest when the evaluation criterion is strict closeness to the observed client-specific minority distribution.

**Downstream Utility for Fraud Detection.** We next examine whether the generated fraud samples improve downstream fraud detection utility. Classifiers are trained on organization-level training sets augmented with one of the candidate sources and evaluated on untouched real organization-level test sets. In line with prior research (Awosika et al., 2024; Özcan et al., 2025) and given the importance of accounting for both false positives and false negatives, we use the F1-score as our primary evaluation metric. As shown in **Table 2**, the overall pattern is clear: our approach achieves the strongest downstream utility across most datasets. In particular, it consistently outperforms the non-augmented baseline and exceeds the performance of SMOTE and ADASYN. The same tendency holds relative to the additional generative baselines in most settings. This performance advantage is not limited to a single augmentation ratio or dataset. Although absolute F1 levels vary across datasets and organization-level partitions, the relative ranking remains largely stable: our approach performs best or among the best across the evaluated settings. The strongest gains are observed for IEEE-CIS, Elliptic, BAF, and PaySim, while results on the Credit Card Fraud dataset remain competitive but less pronounced.

| Datasets | IEEE-CIS | Credit Card Fraud | Elliptic | BAF | PaySim |
|---|---|---|---|---|---|
| **Best Classifier** | **LightGBM** | **HistGradBoosting** | **HistGradBoosting** | **LightGBM** | **Random Forest** |
| Real only (unweighted) | 0.67±0.07 | 0.50±0.22 | 0.93±0.06 | 0.06±0.03 | 0.79±0.15 |
| Real only (weighted) | 0.54±0.13 | 0.52±0.19 | 0.93±0.05 | 0.12±0.01 | 0.79±0.13 |
| Random Oversampling | 0.55±0.14 | 0.72±0.12 | 0.94±0.05 | 0.12±0.01 | 0.78±0.12 |
| SMOTE | 0.66±0.08 | 0.70±0.15 | 0.93±0.05 | 0.21±0.03 | 0.76±0.20 |
| ADASYN | 0.65±0.08 | 0.67±0.16 | 0.92±0.06 | 0.21±0.03 | 0.74±0.24 |
| CTGAN | 0.69±0.06 | **0.77±0.13** | 0.93±0.05 | 0.20±0.04 | 0.81±0.16 |
| TabDDPM | 0.70±0.06 | 0.74±0.10 | 0.93±0.05 | 0.19±0.07 | 0.76±0.13 |
| Local Only DDPM | 0.66±0.06 | 0.72±0.12 | 0.93±0.02 | 0.21±0.03 | 0.81±0.15 |
| Central. DDPM | 0.70±0.06 | 0.71±0.09 | 0.93±0.05 | 0.22±0.03 | 0.81±0.16 |
| **Our Approach** | **0.74±0.05** (***) | 0.76±0.11 | **0.95±0.03** (***) | **0.23±0.02** (***) | **0.82±0.14** (***) |

**Table 2.** Downstream fraud detection performance (F1-score) at the best augmentation ratio for each dataset and best-performing classifier. Bold numbers represent the best performance for each dataset.

## Discussion

Our findings clarify how collaboratively generated synthetic data can create value in privacy-constrained inter-organizational analytics. Returning to our research question, collaborative diffusion does not produce the most locally faithful synthetic fraud samples in terms of distributional proximity. Local interpolation-based methods such as SMOTE and ADASYN often remain closer to observed organization-level fraud distributions. Yet this fidelity advantage does not translate into superior downstream performance. Instead, our approach improves fraud detection more consistently across most datasets, indicating that the most valuable synthetic data are not necessarily those that best reproduce local data geometry, but those that transfer task-relevant structure into local analytical decision processes. This finding speaks directly to the data dependency paradox introduced at the outset. Generative AI methods are often proposed as a way to mitigate limited data availability, but they themselves depend on access to sufficient and diverse high-quality data (Jöhnk et al., 2021; Sharma et al., 2014). In inter-organizational settings, this creates a practical impasse: the data needed to improve AI systems are distributed across organizations, while privacy, regulatory, and governance constraints prevent straightforward pooling (Hirt et al., 2025; Weber et al., 2023). Our results suggest that collaborative synthetic data generation can partially relax this impasse. It does not eliminate the need for data, nor does it provide formal privacy guarantees. However, it offers an analytics architecture through which organizations can benefit from a cross-organizational minority-class structure without centralizing raw transaction records.

A central implication of our results is that synthetic data quality should not be treated as a purely distributional property. In our experiments, fidelity and utility diverge: samples that are statistically closer to local fraud observations are not always the samples that improve fraud detection most. This suggests that fidelity metrics capture only one dimension of synthetic data quality. For organizational analytics, synthetic data also need to be evaluated in terms of downstream task value, governance compatibility, and their ability to improve decision performance under real data-sharing constraints. This distinction is particularly important for data management and analytics research because it challenges the assumption that "better" data are necessarily more locally realistic data. In privacy-constrained analytics, better data may instead be data that make otherwise inaccessible cross-organizational patterns usable for local decision-making.

The results also refine how we understand inter-organizational analytics under data silos. The benefit of our approach does not arise simply because multiple organizations participate in a shared learning process. Rather, the value arises from the specific architecture of collaboration: noisy intermediate representations allow a shared generative process to learn broader fraud-relevant structure, while final generation and downstream model training remain local. In this sense, collaborative synthetic data generation represents a governance-compatible alternative to direct data sharing. It shifts the locus of value creation from raw-data access to structure transfer. For IS research, this is important because it positions synthetic data not merely as an augmentation artifact, but as a mechanism for converting distributed data resources into local analytical and IT capabilities (Barney, 1991; Bharadwaj, 2000). In addition, the findings reveal important boundary conditions. Our approach performs particularly well on IEEE-CIS, Elliptic, BAF, and PaySim, while its advantage is weaker on the Credit Card Fraud dataset. This variation suggests that collaborative synthetic generation is most useful when organizations hold complementary minority-class structures that are not fully observable locally. When fraud patterns are compressed, standardized, or weakly differentiated across organization-level partitions, there is less cross-organizational structure for the collaborative model to transfer. Thus, decentralization alone is not a sufficient condition for value creation. What matters is whether distributed organizations provide analytically complementary information (Jakubik et al., 2024). This boundary condition helps explain why collaborative diffusion is not uniformly dominant and provides a basis for future research on when synthetic data can support inter-organizational analytics.

Our study contributes to data management and analytics research in three ways. First, it extends the evaluation of synthetic data beyond fidelity by showing that local statistical resemblance and downstream decision value can diverge. This calls for multidimensional evaluation frameworks that jointly consider fidelity, utility, governance fit, and privacy risk. Second, it conceptualizes collaborative synthetic data as a mechanism of cross-organizational knowledge transfer. Rather than requiring organizations to share raw records, collaborative generation can encode useful structure from distributed data sources and make that structure available for local analytics. Third, it highlights the role of analytics architectures in shaping AI performance. The split between shared and local denoising is not only a technical design choice; it reflects an organizational governance arrangement about what can be learned jointly for developing shared generative AI resources and what must remain under local control.

For organizations, the findings suggest that synthetic data initiatives should not be selected based on generative realism alone. A method that closely reproduces local records may be less valuable than one that improves the downstream decision task. Financial institutions and other organizations operating under strict data-sharing constraints should therefore evaluate synthetic data methods within the full analytics pipeline: generation, augmentation, model training, and decision performance. They should also compare collaborative generative approaches against simpler baselines such as class weighting, random oversampling, SMOTE, and ADASYN. From a resource perspective, the timestep-split architecture also changes what participation costs an organization. Rather than provisioning the infrastructure to train and operate a complete generative model locally, a participating organization executes only the final share of the denoising trajectory and exchanges batch-sized noised representations instead of model parameters (Allmendinger et al., 2026). At the same time, organizations should not interpret collaborative generation as a substitute for privacy analysis. Because synthetic data can still leak information about original distributions, deployment should be accompanied by privacy testing, governance controls, and clear rules about what information is exchanged between clients and shared infrastructure.

### *Limitations*

This study has several limitations. First, while we compare collaborative diffusion with multiple local, centralized, and non-generative baselines, we do not fully benchmark it against alternative collaborative generative paradigms beyond the tested federated variants. In particular, our efficiency argument is architectural. We quantify the organization-side workload through the share of denoising steps executed locally and the size of the exchanged representations, but we do not report empirical wall-clock training times, memory footprints, or communication volumes. Second, the number of clients is limited, and larger client populations may introduce stronger heterogeneity and coordination challenges. Third, the findings are specific to tabular fraud data and should not be generalized directly to other modalities. Finally, synthetic data quality does not guarantee privacy, as useful synthetic samples may still leak information about the original distribution (Stadler et al., 2022). Thus, our contribution concerns a privacy-preserving collaboration architecture under realistic data-sharing constraints rather than formal privacy guarantees.

## Conclusion and Outlook

This paper examines collaborative diffusion for fraud detection as an instance of a broader IS problem: organizations need to create analytical value from data that are rare, sensitive, fragmented, and difficult to share across organizational boundaries. Building on resource-based views and IT capability (Barney, 1991; Bharadwaj, 2000), we argue that the strategic value of data depends not only on ownership, but also on analytics capabilities that enable firms to mobilize and recombine distributed data resources for local decision-making (Hirt et al., 2025). Collaborative diffusion can therefore be understood as an IT-enabled capability transforming constrained inter-organizational resources into decision-relevant analytical value.

Our findings show that collaborative synthetic data generation can improve local fraud detection by transferring task-relevant structures across organizational boundaries without pooling raw records. This extends analytics research linking data use to organizational decision-making (Sharma et al., 2014) and aligns with data-centric AI research that emphasizes the design and mobilization of useful training data rather than model optimization alone (Jakubik et al., 2024). For IS research, the key implication is that synthetic data should not be evaluated only by distributional fidelity, but also by downstream utility, governance compatibility, and its role in inter-organizational value creation. At the same time, collaborative diffusion raises governance questions central to IS research. Synthetic representations may leak sensitive patterns, misrepresent local contexts, or obscure the value of individual data contributions. Future research should therefore examine safeguards for privacy, usage rights, accountability, and compensation in collaborative synthetic data ecosystems. In line with emerging data valuation research in federated settings (Han et al., 2026), future work should study when structure transfer creates value rather than negative transfer, how rare or minority-class contributions can be valued, and how synthetic data artifacts should be governed across organizational boundaries.